\documentclass[nofootinbib,superscriptaddress,twocolumn,aps,pra]{revtex4-1}

\usepackage{amsmath,amsfonts,amssymb}
\usepackage{wrapfig}
\usepackage{graphicx}
\usepackage{bbm}
\usepackage{color}
\usepackage{float}
\usepackage{subfigure}
\usepackage{textcomp}
\usepackage[english]{babel}
\usepackage{pifont}

\usepackage[colorlinks=true,linkcolor=blue,citecolor=blue]{hyperref}
\newcommand{\nn}{\nonumber}
\newcommand{\djj}{d\kern-0.4em\char"16\kern-0.1em}

\makeatletter

\def\compoundrel#1\over#2{\mathpalette\compoundreL{{#1}\over{#2}}}
\def\compoundreL#1#2{\compoundREL#1#2}
\def\compoundREL#1#2\over#3{\mathrel
      {\vcenter{\hbox{$\m@th\buildrel{#1#2}\over{#1#3}$}}}}
\makeatother

\usepackage{babel}
\usepackage{wasysym}
\usepackage{harmony}
\usepackage{semtrans}

\makeatother

\def\be{\begin{eqnarray}}
\def\ee{\end{eqnarray}}
\def\bee{\begin{eqnarray*}}
\def\eee{\end{eqnarray*}}

\begin{document}

\title{Meta-learning within Projective Simulation}

\author{Adi Makmal}
\affiliation{Institut f{\"u}r Theoretische Physik,
Universit{\"a}t Innsbruck, Technikerstra{\ss }e 21a, A-6020 Innsbruck, Austria}
%\affiliation{Institut f{\"u}r Quantenoptik und Quanteninformation der
%\"Osterreichischen Akademie der Wissenschaften, Innsbruck, Austria}
\author{Alexey A. Melnikov}
\affiliation{Institut f{\"u}r Theoretische Physik,
Universit{\"a}t Innsbruck, Technikerstra{\ss }e 21a, A-6020 Innsbruck, Austria}
\affiliation{Institut f{\"u}r Quantenoptik und Quanteninformation der
\"Osterreichischen Akademie der Wissenschaften, Technikerstra{\ss }e 21a, A-6020 Innsbruck, Austria}
\author{Vedran Dunjko}
\affiliation{Institut f{\"u}r Theoretische Physik,
Universit{\"a}t Innsbruck, Technikerstra{\ss }e 21a, A-6020 Innsbruck, Austria}
%\affiliation{Institut f{\"u}r Quantenoptik und Quanteninformation der
%\"Osterreichischen Akademie der Wissenschaften, Innsbruck, Austria}
%\affiliation{Laboratory of Evolutionary Genetics, Division of Molecular Biology, Ru\djj er Bo\v{s}kovi\'{c} Institute, Bijeni\v{c}ka cesta 54, 10000 Zagreb, Croatia.} 
\author{Hans J. Briegel}
\affiliation{Institut f{\"u}r Theoretische Physik,
Universit{\"a}t Innsbruck, Technikerstra{\ss }e 21a, A-6020 Innsbruck, Austria}

\date{\today}
\begin{abstract}
% internal parameters in AI models
% ML
% insert ML to PS
% General, unifoem, design, that builds upon the basic design of the PS
Learning models of artificial intelligence can nowadays perform very well on a large variety of tasks. 
However, in practice different task environments are best handled by different learning models, rather than a single, universal, approach.
%Yet, there exist no single, universal, model that can cope with all or even the majority of canonical tasks. 
Most non-trivial models thus require the adjustment of several to many learning parameters, which is often done on a case-by-case basis by an external party. Meta-learning refers to the ability of an agent to autonomously and dynamically adjust its own learning parameters, or meta-parameters. 
In this work we show how projective simulation, a recently developed model of artificial intelligence, can naturally be extended to account for meta-learning in reinforcement learning settings. 
The projective simulation approach is based on a random walk process over a network of clips. The suggested meta-learning scheme builds upon the same design and employs clip networks to monitor the agent's performance and to adjust its meta-parameters ``on the fly". 
We distinguish between ``reflexive adaptation" and ``adaptation through learning", and show the utility of both approaches. In addition, a trade-off between flexibility and learning-time is addressed. The extended model is examined on three different kinds of reinforcement learning tasks, in which the agent has different optimal values of the meta-parameters, and is shown to perform well, reaching near-optimal to optimal success rates in all of them, without ever needing to manually adjust any meta-parameter. 
%This holds also when the external environment suddenly changes. 
%In this work we extend the projective simulation, a recently developed model of artificial intelligence, with a meta-learning component. We show how the basic design of the projective simulation model 
%
%In this work we develop a component of meta-learning into projective simulation, a recently developed model of artificial intelligence, to 
% (e.g.\ $\gamma$, $\eta$, etc.) to optimal values with respect to its present scenario. For example, we have already seen that a very small value of the damping parameter $\gamma$ facilitates high learning efficiencies, but that at the same time it also sabotages the agent's ability to adapt to changing environments. Currently, the value of $\gamma$ is defined externally, but ideally, a good balance between these contradicting qualities would be reached 
%by the agent itself through its experience. 
%While meta-learning is a beneficial concept, an elegant and economical implementation of it is not trivial. In particular one should ensure that the scheme does not blow up the model by introducing new parameters that would also need to be adjusted. Here a first small step toward reaching this goal is described.
\end{abstract}

% keywords: meta-learning, meta-knowledge

\maketitle

\section{Introduction}
% what is meta learning
	There are many different kinds of artificial intelligent (AI) schemes. These schemes differ in their design, purpose, and underlying principles~\cite{AI_ModernApproach}. 
One feature common to all non-trivial proposals is the existence of learning parameters, which reflect certain assumptions or bias about the task or environment with which the agent has to cope \cite{2009_Metalearning_book}. Moreover, as a consequence of the so-called no-free lunch theorems~\cite{NFL97}, it is known that it is impossible to have a fixed set of parameters %(a parameter-free model) 
which are optimal for all task environments. In practice these parameters (which, for some schemes, may be more than a dozen, e.g.\ in the extended learning classifier systems  \cite{wilson1995classifier}) are typically fine-tuned manually by an external party (the user), on a case-by-case basis.
%, for obtaining optimal performance. 
An autonomous agent, however, is expected to adjust its learning parameters, automatically, by itself. Such a self monitoring and adaptation of the agent's own internal settings is often termed as \emph{meta-learning}~\cite{2009_Metalearning_book, schaul2010metalearning, 2004_ML_Carrier}.

In the AI literature the term meta-learning, defined as ``learning to learn"~\cite{thrun1998learning} or as a process of acquiring meta-knowledge~\cite{2004_ML_Carrier, 2009_Metalearning_book}, is used in a broad sense and accounts for various concepts. These concepts are tightly linked to practical problems, two of which are mostly considered in the context of meta-learning. In the first problem meta-learning accounts for a selection of a suitable learning model for a given task~\cite{2002_Duch_InBook, brazdil2003ranking, zhao2006model, adankon2009model, AbdulrahmanEtAl:MetaSel2015} or combination of models~\cite{todorovski2003combining}, including automatic adjustments when the task is changed. In the second problem meta-learning accounts for automatic tuning of model learning parameters, also referred to as meta-parameters~\cite{schaul2010metalearning, Ishii2002665, 2003_Schweighofer_ML_in_RL, eriksson2003evolution, kobayashi2009, tokic2013} in reinforcement learning (RL) or hyperparameters~\cite{bengio2000gradient, bardenet2010surrogating, bergstra2012random, reif2012meta, Thornton:2013:ACS, smith2014recommending, feurer2015initializing} in supervised learning.

Both of these concepts of meta-learning are widely addressed in the framework of supervised learning. The problem of supervised learning model selection, or algorithm recommendation, is solved by, e.g., the k-Nearest Neighbor algorithm~\cite{brazdil2003ranking}, similarity-based methods~\cite{2002_Duch_InBook}, meta decision trees~\cite{todorovski2003combining} or an empirical error criterion~\cite{adankon2009model}. The second problem in meta-learning, the tuning of hyperparameters, is usually solved by gradient-based optimization~\cite{bengio2000gradient}, grid search~\cite{bardenet2010surrogating}, random search~\cite{bergstra2012random}, genetic algorithms~\cite{reif2012meta} or Bayesian optimization~\cite{Thornton:2013:ACS, feurer2015initializing}. 
Approaches to combined algorithm selection and hyperparameter optimization were recently presented in Refs.~\cite{Thornton:2013:ACS, smith2014recommending, feurer2015initializing}.

In the RL framework, where an agent learns from interacting with a rewarding environment~\cite{SuttonBarto98}, the notion of meta-learning usually addresses the second practical problem, the need to automatically adjust meta-parameters, such as the discount factor, the learning rate and the exploitationÂ-exploration parameter~\cite{Ishii2002665, 2003_Schweighofer_ML_in_RL, eriksson2003evolution, Achbany20082507, kobayashi2009, tokic2013}. In the context of RL it is also worthwhile to mention the G{\"o}del machine~\cite{2005_Schmidhuber_Godel_machine}, which is, due to its complexity, of interest predominantly as a theoretical construction in which all possible meta-levels of learning are contained in fully self-referential learning system.

%Meta-learning is still in the centre of attention 

%maybe list out examples: eg. alpha in Q-learning, gamma in PS, balba, but, more abstractly, also the choice of the activation function in neural networksÂ
% meta-learning in the literature (AI -> ML -> RL)

In this paper we develop a simple form of meta-learning for the recently introduced model of projective simulation (PS) \cite{2012_Briegel_PSI}. 
The PS is a model of artificial intelligence that is particularly suited to RL problems (see \cite{2013_Mautner_PSII,2014_Melnikov_PSIII,bjerland2015projective} where the PS was shown to perform well, in comparison to more standard RL machinery, on both toy- and real-world tasks such as the ``grid-world", the ``mountain-car", the ``cart-pole balancing" problem and the ``Infinite Mario" game, and see \cite{2015_Melnikov_PS_Generalization} where it handles infinitely large RL environments through a particular generalization mechanism). The model is physics-oriented, aiming at an embodied \cite{Understanding_Intelligence_1999} (rather than computational) realization, with a random-walk through its memory structure as its primary process. PS is based on a special type of memory, called the \emph{episodic \& compositional memory} (ECM), that can be represented as a directed weighted graph of basic building blocks, called \emph{clips}, where each clip represents a memorized percept, action, or combinations thereof. Once a percept is perceived by the PS agent, the corresponding percept clip is activated, initiating a random-walk on the clip-network, that is the ECM, until an action clip is hit and the corresponding action is performed by the agent. This realizes a stochastic processing of the agent's experience.  

The elementary process of the PS, namely the random-walk, is an established theoretical concept, with known applications in randomized algorithms \cite{Randomized_Algorithms_1995}, thus providing a large theoretical tool box for designing and analyzing the model. Moreover, the random walk can be extended to the quantum regime, leading to \emph{quantum walks} \cite{1965_Feynman, 1993_Aharonov, Aharonov:2001}, in which case polynomial and even exponential improvements have been reported in e.g. hitting and mixing times \cite{Childs:2003, Kempe2005, KMOR}. The results in the theory of quantum walks suggest that improvements in the performance of the PS may be achievable by employing these quantum analogues. Recently, a quantum variant of the PS (envisioned already in \cite{2012_Briegel_PSI}) was indeed formalized and shown to exhibit a quadratic speed-up in deliberation time over its classical counterpart \cite{2014_Paparo_QPS, dunjko2015quantum, friis2015coherent}. 
  
From the perspective of meta-learning, the PS is a comparatively simple model with few  number of learning parameters \cite{2013_Mautner_PSII}. This suggests that providing the PS agent with a meta-learning mechanism may be done while maintaining its overall simplicity. In addition to simplicity, we also aim at structural homogeneity: the meta-learning component should be combined with the basic model in a natural way, with minimal external machinery. 
In accordance with these requirements, the meta-learning capability which we develop here is based on supplementing the basic ECM network, which we call the base-level ECM network, with additional meta-level ECM networks that dynamically monitor and control the PS meta-parameters. This extends the structure of the PS model from a single network to several networks that influence each other. 

In general, when facing the challenge of meta-learning in RL, one immediately encounters a trade-off between efficiency (in terms of learning times) and success rates (in terms of achievable rewards), on the one side, and flexibility on the other side (as pointed out, e.g.,\ also in \cite{2007_AI_Magazine_Anderson}). 
Humans, for example, are extremely flexible and robust to changes in the environment, but are not very efficient and reach sub-optimal success rates. Machines, on the other hand, can learn fast and perform optimally at a given task (or a family of tasks), yet fail completely on another. Clearly, to achieve a level of robustness, machines would have to repeatedly revise and update their internal design, i.e.\ to meta-learn, a process which necessarily takes time. Moreover, reaching optimal success rates in certain tasks, may require an over-fitting of the scheme's meta-parameters, which might harm its success in other tasks. It can therefore be expected that any form of meta-learning (which improves the flexibility of the model), may do so at the expense of the model's efficiency and (possibly even) success rates, and we will observe this inclination also in our work. 

Another aspect of meta-learning which we highlight throughout the paper is the underlying principles that govern the agent's internal adjustment. Here, we distinguish between two different (sometimes complementary) alternatives which we call \emph{reflexive adaptation} and \emph{adaptation through learning}. 
Informally, by reflexive adaptation we mean that the agent's meta-parameters are adjusted via a fixed recipe (which may or may not be deterministic), which takes into account only the recent performance of the agent, while ignoring the rest of the agent's history. Essentially, this amounts to adaptation without a need for additional memory. An example for such a reflexive adaptation approach for meta-learning can be found in \cite{2003_Schweighofer_ML_in_RL} where the fundamental RL parameters, namely, the learning rate $\alpha$, the exploitation-exploration parameter $\beta$, and the discount factor $\gamma$ are tuned according to predefined equations; In contrast, an agent which adapts its parameter through learning, exploits to that end its entire individual experience. Accordingly, adaptation through learning does require an additional memory. In this work we consider both kinds of approaches\footnote{The terminology we employ is based on the basic classification of intelligent agents; if we perceive the meta-learning machinery as an agent, then the reflexive adaptation mechanism corresponds to simple reflexive agents, whereas the learning adaptation mechanism corresponds to a learning agent.}.

The paper is structured as follows: Section \ref{sec:PS_model} shortly describes the PS model including its meta-parameters. Section \ref{sec:Necessity_ML_PS} demonstrates the advantages of meta-learning, by considering explicit task scenarios where the PS model has different optimal values of the meta-parameters. In Section \ref{sec:ML_in_PS_design} we present the proposed meta-learning design and explain how it combines with the basic model. The model is then examined and analyzed through simulations in Section \ref{sec:results}, where the performance of the meta-learning PS agent is evaluated in three different types of changing environments. Throughout this section the proposed meta-learning scheme is further compared to other, more naive, alternatives of meta-learning schemes. Finally, Section \ref{sec:discussion} concludes the paper and discusses some of its open questions.  

\section{The PS model} \label{sec:PS_model}
For the benefit of the reader we first give a short summary of the PS; for a more detailed description, including recent developments see~\cite{2012_Briegel_PSI, 2013_Mautner_PSII, 2014_Melnikov_PSIII, 2015_Melnikov_PS_Generalization, bjerland2015projective}. 
The central component of the PS model is the episodic $\&$ compositional memory (ECM), formally a network of \emph{clips}. 
The possible clips include percept clips (representing a percept) and action clips (representing an action), but can also include the representations of various combinations of percept and action sequences (thus representing e.g. an elapsed exchange between the agent and environment, or subsets of the percept space as occurring in the model of PS with generalization \cite{2015_Melnikov_PS_Generalization}).
Within the ECM, a clip $c_i$ may be connected to clip $c_j$ via a weighted directed edge, with a corresponding time-dependent real positive weight $h^{(t)}(c_i, c_j)$ (called $h$-value), which is larger than or equal to its initial value of $h_0 = 1$. 

The deliberation process of the agent corresponds to a random walk in the ECM, where the transition probabilities are proportional to the  $h-$values. More specifically, upon encountering a percept, the clip corresponding to that percept is activated, and a random walk is initiated. The transition probability from clip $c_i$ to $c_j$ at time step $t$, corresponds to the re-normalized $h$-values:
\begin{equation}
  p^{(t)}(c_j|c_i) = \dfrac{h^{(t)}(c_i,c_j)}{\sum_{k} h^{(t)}(c_i,c_k)}.
\label{eq:probablititesBasic}
\end{equation}
The random walk is continued until an action clip has been hit, upon which point the corresponding action is carried out.

The learning aspect of the PS agent is achieved by the dynamic modification of the $h-$values, depending on the response of the environment. 
Formally, at each time-step, the $h$-values of the edges that were traversed during the preceding random walk are updated as follows:
\begin{equation}
  h^{(t+1)}(c_i,c_j) = h^{(t)}(c_i,c_j) - \gamma (h^{(t)}(c_i,c_j) - 1) + \lambda,
\label{eq:hupdate}
\end{equation}
where $0\leq\gamma\leq 1$ is a damping parameter and $\lambda$ is a non-negative reward given by the environment. 
The $h$-values of the edges which were not traversed during the preceding random walk are not rewarded (no addition of $\lambda$), but are nonetheless damped away toward their initial value $h_0 = 1$ (by the $\gamma$ term).
%In the expression above, the introduced the superscript ($t$, $t+1$) specifies the time-step, and all initial $h-$values are set to 1, so $h^{(0)}=1$.
With this update rule in place, the probability to take rewarded actions is increased with time, that is, the agent learns.

The damping parameter $\gamma$ is a meta-parameter of the PS model. The higher it is, the faster the agent forgets its knowledge. For certain settings, introducing additional parameters to the ECM network can lead to better learning performance. A particularly useful generalization is the ``edge glow'' mechanism, introduced to the model in~\cite{2013_Mautner_PSII}. 
Here, an additional time-dependent variable $0\leq g \leq 1$ is attributed to each edge of the ECM, and a term depending on its value is added to the update rule of the $h-$values:
\begin{eqnarray}
  h^{(t+1)}(c_i,c_j) &=& h^{(t)}(c_i,c_j) - \gamma (h^{(t)}(c_i,c_j) - 1) \nn \\ 
 &+& g^{(t)}(c_i,c_j)\lambda.
\label{eq:hupdate2}
\end{eqnarray}
This update rule holds for all edges, so that edges which were not traversed still may end up being enhanced, proportional to their $g-$value. 
The $g-$value dynamically changes. Each time an edge is traversed, its $g$-value is set to $g=1$, and dissipates in the following time steps with a rate $\eta$:
\begin{equation}
  g^{(t+1)}(c_i,c_j) = g^{(t)}(c_i,c_j)(1 - \eta).
\label{eq:gupdate}
\end{equation}
The $\eta$ parameter is thus another meta-parameter of the model. %, which is denoted as the ``glow parameter'',

The decay of the $g$-values ensures that the reward effects the edges traversed at different points in time, to a different extent. In particular, recently traversed edges are enhanced more (after a rewarding step), relative to edges traversed in the more remote past. The $\eta$ parameter controls the strength of this temporal dependence. For instance, a low value of $\eta$ implies that the edges which were traversed a while back in the past will nonetheless be enhanced. In contrast, by setting $\eta=1$, only the last traversed path is enhanced in which case the update rule reverts back to Eq.~(\ref{eq:hupdate}).

The glow mechanism thus establishes temporal correlations between percept-action pairs, and enables the agent to perform well also in settings where the reward is delayed (e.g. in the grid-world and the mountain-car tasks~\cite{2014_Melnikov_PSIII}) and/or contingent on more than just the immediate history of agent-environment interaction (such as in the $n$-ship game, as presented in \cite{2013_Mautner_PSII}). 
%This was illustrated in Ref.\ \cite{2013_Mautner_PSII} for the $n$-ship problem, in which the environment rewards the agent small problems in which the  and in Ref.\ \cite{2014_Melnikov_PSIII} for the grid world and the mountain car problem, which are real-world tasks 
%Such environment are encountered, for example, in the grid-world and the mountain car problem (such esee \cite{2013_Mautner_PSII} and \cite{2014_Melnikov_PSIII}).

The basic variant of the PS model (so-called two-layered variant) can be formally contrasted to more standard RL schemes, where it closely resembles the SARSA algorithm~\cite{rummery1994line}. An initial analysis of the relationship of the two models was given in \cite{2015_Melnikov_PS_Generalization}. Readers familiar with the SARSA model may benefit from the observation that the functional roles of the $\alpha$ and $\gamma$ parameters in SARSA are closely matched by the $\gamma$ and $\eta$ parameters of the PS, respectively. However, as the PS is explicitly not a state-action value function-based model, the analogy is not exact. For more details we refer the interested reader to \cite{2015_Melnikov_PS_Generalization}. In the following section, we describe the behaviour of the PS model, and the functional role of its meta-parameters in greater detail.

\section{Advantages of Meta-learning} \label{sec:Necessity_ML_PS}
The basic memory update mechanism of the PS, as captured by Eq.~(\ref{eq:hupdate2})-(\ref{eq:gupdate}) has two meta-parameters, namely the damping parameter $\gamma$, and the glow parameter $\eta$. In what follows, we examine the role of these parameters in the learning process of the agent. We then demonstrate, through examples, that for none of these parameters there is a unique value that is universally optimal, i.e.\ that different environments induce different optimal $\gamma$ and $\eta$ values. These examples provide direct motivation for introducing a meta-learning mechanism for the PS model.

%These examples also provide direct motivation for the introduction of mechanism for the dynamic adjustment of the parameters influenced by experience (or history of the agent? I think history is more standard in RL) ; comment: this alternative sentence already partially admits that our mechanisms are tailored. Which is ok, and I think we should be, to this level, clear about this.

	\subsection{Damping: the $\gamma$ parameter} \label{secsec:damping_parameter}
	The damping parameter $0\leq\gamma\leq 1$ controls the forgetfulness of the agent, by continuously damping the $h$-values of the clip network. 
A direct consequence of this is that a non-zero $\gamma$ value bounds the $h$-values of the clip network to a finite value, which in turn limits the maximum achievable success probability of the agent. 
As a result, in many typical tasks considered in the RL literature (grid-world, mountain-car, and tic-tac-toe, to name a few), in which the environments are consistent, i.e.\ not changing, the optimal performance is achieved without any damping, that is by setting $\gamma=0$. 

However, when the environment does change, the agent may need to modify its ``action pattern", which implies varying the relative weights of the $h$-values. Presetting a finite $\gamma$ parameter would then quicken the agent's learning time in the new environment, at the expense of reaching lower success probabilities, as demonstrated and discussed in Ref.\ \cite{2012_Briegel_PSI}. This gives rise to a clear trend: The higher the value of $\gamma$, the faster is the relearning in a changing environment, and the lower is the agent's asymptotic success probability.

\begin{figure}[h]
	\begin{center}
		%\begin{minipage}{9cm}
				\includegraphics[width=8cm]{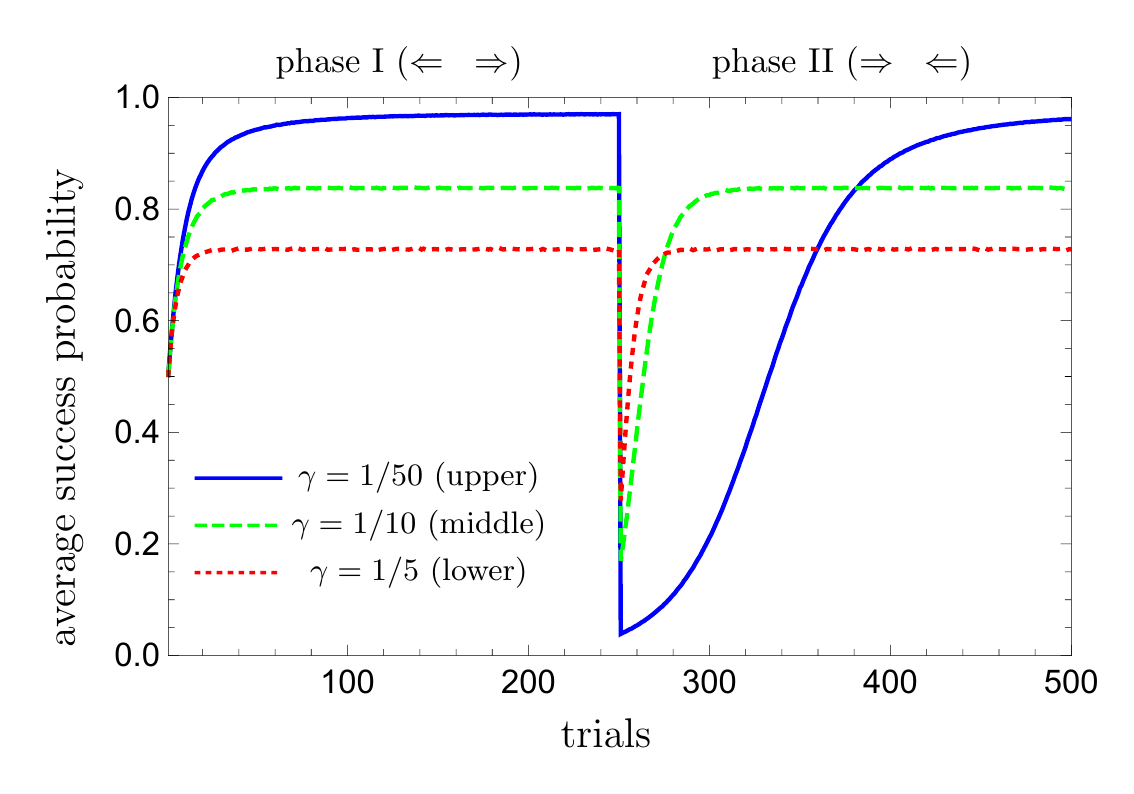}
			%\end{minipage}
		\end{center}
	\caption{(Color online) \emph{Invasion game:} The attacker inverts its strategy after 250 steps. The agent's average success probability is plotted as a function of number of trials (games). A trade-off between success probability and relearning time is depicted for different $\gamma$ values. An optimal value of $\eta=1$ is used. The simulation was done by averaging over $10^6$ agents. Adapted from \cite{2012_Briegel_PSI}.}
	\label{fig:invasion_game_gamma_tradeoff}
\end{figure}

The trade-off between learning time and success probability in changing environments can be demonstrated on the invasion game~\cite{2012_Briegel_PSI} example. The invasion game is a special case of the contextual multi-armed bandit problem~\cite{Bandits05} and has no temporal dependence. In this game an agent is a defender and should try to block an attacker by moving in the same direction (left or right) with the attacker. Before making a move, the attacker shows a symbol (``$\Leftarrow$" or ``$\Rightarrow$"), which encodes its future direction of movement. Essentially, the agent has to learn where to go for every given direction symbol. Fig.~\ref{fig:invasion_game_gamma_tradeoff} illustrates how the PS agent learns by receiving rewards for successfully blocking the attacker. Here, during a phase of 250 steps, the attacker goes right (left) whenever it shows a right (left) symbol, but then, at the second phase of the game, the attacker inverts its rules, and goes right (left) whenever it shows left (right). It is seen that higher values of $\gamma$ yield lower success probabilities, but allow for a faster learning in the second phase of the game. 

%\footnote{The number of agents we use in our simulations varies from one task to another, according to the method we use to calculate the agent's performance: averaging over the agents' success probabilities requires many agents to obtain a meaningful statistics, whereas when for each agent the average success probability is directly accessed through its network, much fewer agents are needed (since the average is then performed over averages). See also~\cite{2013_Mautner_PSII} for a detailed discussion on error bars in such simulations.}

\begin{figure}[h!]
	\begin{center}
		\begin{minipage}{9cm}				\includegraphics[width=8cm]{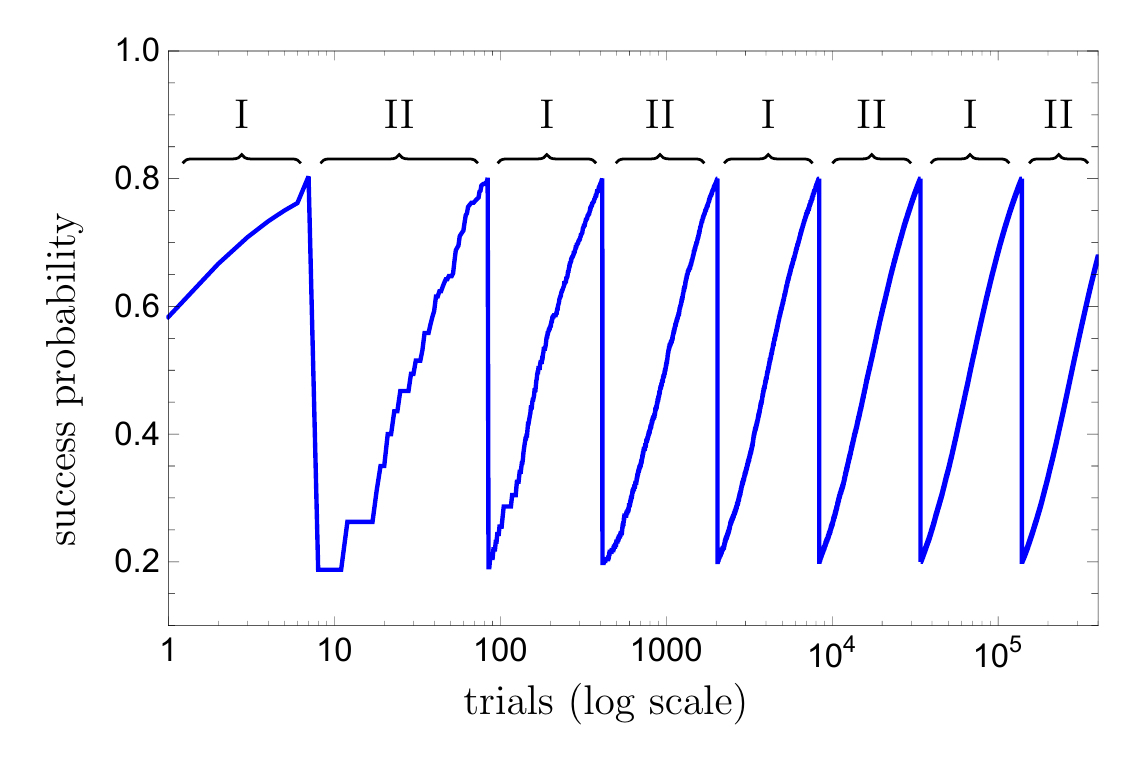}
			\end{minipage}
		\end{center}
	\caption{\emph{Invasion game:} The attacker inverts its strategy whenever the agent's success probability reaches 0.8. The agent's performance is plotted as a function of number of trials on a log scale, demonstrating learning times that increase exponentially with the number of inversions. The simulation was done with a single agent, where the success probabilities were extracted directly from the agent's base-level ECM network. Meta-parameters: $\gamma=0$, $\eta=1$.}
	\label{fig:invasion_game_change_per_suc}
\end{figure}

To illustrate further the slow-down of the learning time in a changing environment when setting $\gamma = 0$, Fig.~\ref{fig:invasion_game_change_per_suc} shows the average success probability of the basic PS agent in the invasion game as a function of number of trials, on a log scale. Here the attacker inverts its rules whenever the agent reaches a certain success probability (here set to 0.8). We can see that the time that the agent needs to learn at each phase grows exponentially in the number of the changes of the phases, requiring more and more time for the agent to learn, so that eventually, for any finite learning time, there will be a phase for which the agent fails to learn. 

\begin{figure}[h!]
	\begin{center}
		\begin{minipage}{9cm}
				\includegraphics[width=8cm]{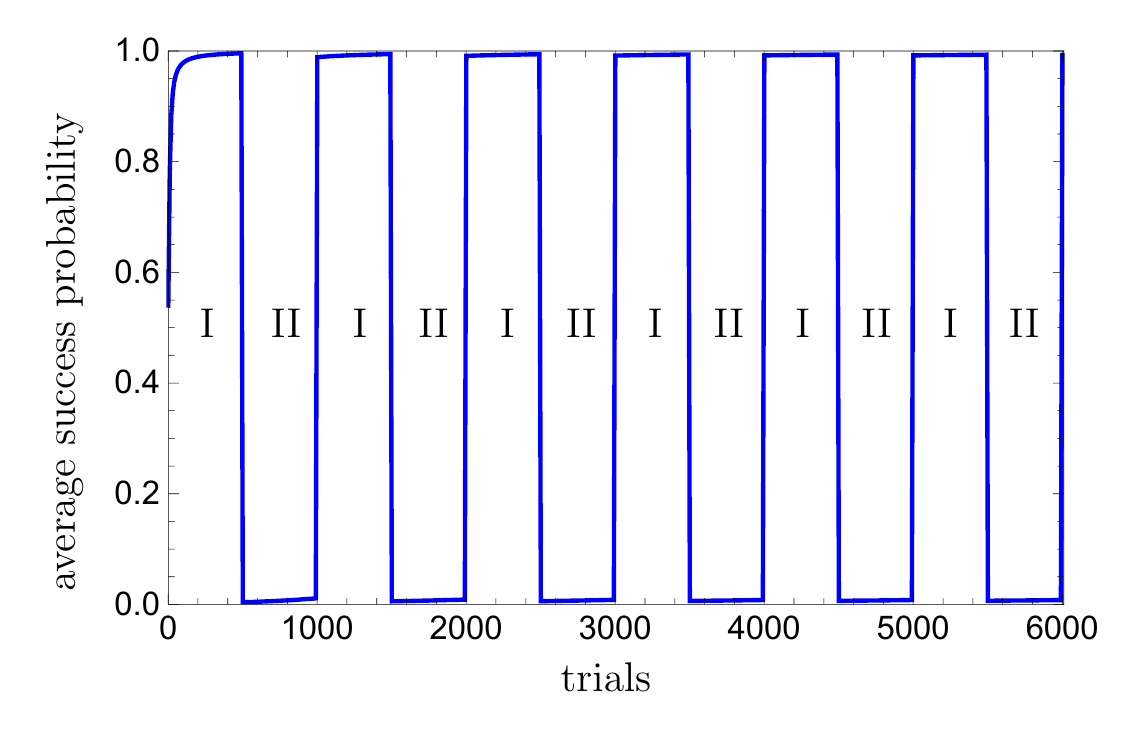}
			\end{minipage}
		\end{center}
	\caption{\emph{Invasion game:} The attacker changes its strategy every 500 steps. The agent's average success probability is plotted as a function of number of trials, demonstrating that only one of the two set of the attacker's strategy can be learned. Moreover, the performance of the agent, averaged over the two phases, converges to the performance of a random agent.
The simulation was done by averaging over 100 agents, where for each agent the success probabilities were extracted directly from its base-level ECM network. Meta-parameters: $\gamma=0$, $\eta=1$.}
	\label{fig:invasion_game_fixed_change}
\end{figure}

Setting a zero damping parameter in changing environments may even be more harmful for the agent than merely increasing its learning time. To give an example, consider an invasion game, where the attacker inverts its rules every fixed finite number of steps. Without damping, the agent will only be able to learn a single set of rules, while utterly failing on the inverted set. This is shown in Fig.~\ref{fig:invasion_game_fixed_change}.
% illustrating how using a zero damping parameter can drastically sabotage the agent's performance in changing environments.  

The performance of the PS agent in the considered examples, shown in Figs.~\ref{fig:invasion_game_gamma_tradeoff} -- \ref{fig:invasion_game_fixed_change}, suggests that it is important to raise the $\gamma$ parameter whenever the environment changes (the performance drops down) and set it to zero whenever the performance is steady. As we will show in Section \ref{secsec:ECM_gamma} this adjustment can be implemented by means of reflexive adaptation. However the reflexive adaptation of the $\gamma$ parameter makes meta-learning less general, because it fixes the rule of the parameter adjustment. To make the PS agent more general we will implement $\gamma$ adaptation also through learning, which gives the agent the possibility to \textit{learn} the opposite rule, i.e. to decrease $\gamma$ whenever the agent's performance goes down.

So far we assumed that the glow mechanism is turned off by setting $\eta=1$, which is optimal for the invasion game. The same holds in all environments where the rewards depend only on the current percept-action pair, with no temporal correlations to previous percepts and actions. In the next section, however, we look further into scenarios where such temporal correlations do exist, and study their influence on the optimal $\eta$ value.

\subsection{Glow: the $\eta$ parameter} \label{secsec:glow_parameter}
In task environments where reward from an environment is a consequence of a series of decisions made by an agent, it is vital to ensure that not only the last action is rewarded, but the entire sequence of actions. Otherwise, these previous actions, which eventually led to a rewarded decision, will not be learned. 
As described in Sec.~\ref{sec:PS_model}, rewarding a sequence of actions is done in the PS model by attributing a time dependent $g$-value to each edge of the clip network and rewarding the edge with a reward proportional to its $g$-value. Once an edge is excited, its $g$-value is set to $g=1$, whereas all other $g$-values decay with a rate $\eta$, which essentially determines the extent to which past actions are rewarded. As we show next, the actual value of the $\eta$ parameter plays a crucial role in obtaining high average reward, its optimal value depends on the task, and finding it is not trivial.

Here we study the role of the $\eta$ parameter in the $n$-ship game example, introduced in~\cite{2013_Mautner_PSII}. In this game $n$ ships arrive in a sequence, one by one, and the agent is capable of blocking them. If the agent blocks one or several ships out of the first $n-1$ ships, it will get a reward of $\lambda_\mathrm{min}=1$ for each ship immediately after blocking it. If, however, the agent will refrain from blocking the ships, although there is an immediate reward for that, it will get a larger reward of $\lambda_\mathrm{max}=5\times(n-1)$ for blocking only the last, $n$-th ship. In this scenario the optimal strategy differs from the greedy strategy of collecting immediate rewards, because the reward $\lambda_\mathrm{max}$ is larger than the sum of all the small rewards that can be obtained during the game.

%In the n-ship game a ship arrives and the agent should decide whether to block it or not. After an action of the agent the next ship comes and the agent should decide again. This process repeats until the agent decide upon the last, $n$-th ship. The goal of the task is to skip the first $n-1$ ships and to block the last one. For this sequence of actions the agent will get a large reward $\lambda_\mathrm{max}=5\times(n-1)$ at the end of the game. However if the agent blocks one or several ships out of the first $n-1$ ships, it will get a smaller reward $\lambda_\mathrm{min}=1$ after blocking each ship, but will not get a large reward at the end of the game.

\begin{figure}[h]
	\begin{center}
		\begin{minipage}{9cm}
				\includegraphics[width=8cm]{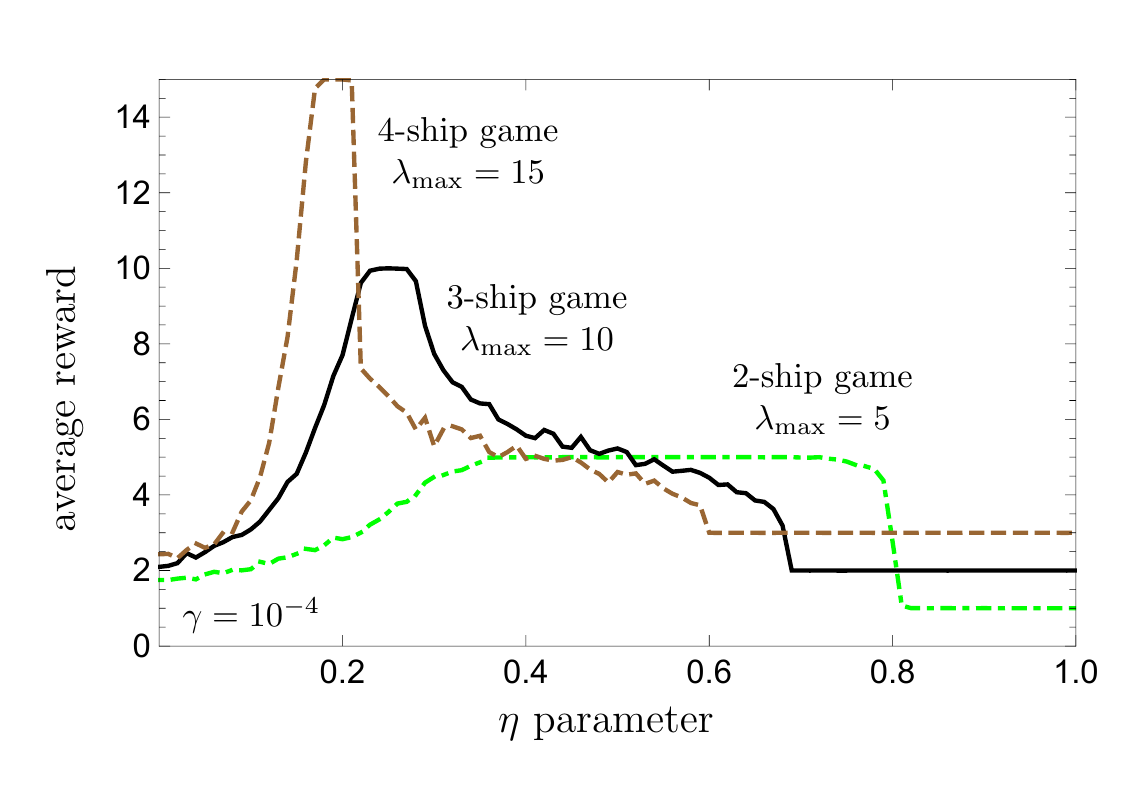}
			\end{minipage}
		\end{center}
	\caption{(Color online) \emph{$n$-ship game:} The dependence of the performance on the $\eta$ parameter is shown for different $n$. The performance is evaluated by an average reward gained during the $10^6$-th game. The simulation was done by averaging over $10^3$ agents. The $\gamma$ parameter was set to $10^{-4}$. Adapted from \cite{2013_Mautner_PSII}.}
	\label{fig:nship_eta_reconstructed}
\end{figure}

The optimal strategy in the described game can be learned by using the glow mechanism and by carefully choosing the $\eta$ parameter. The optimal $\eta$ value depends on the number of ships $n$, as shown in Fig.~\ref{fig:nship_eta_reconstructed}, where the dependence of the average reward received during the game on the $\eta$ parameter is plotted for each $n \in\{2,3,4\}$. It is seen that as the number of $n$ ships grows, the best average reward is obtained using a smaller $\eta$ value, i.e. the optimal $\eta$ value decreases. This makes sense as a smaller $\eta$ value leads to larger sequences of rewarded actions.

%-------------

The simulations of the $n$-ship game shown in Fig.~\ref{fig:nship_eta_reconstructed} emphasize the importance of setting a suitable $\eta$ value. In other, more involved scenarios, such a dependency may be more elusive, making the task of setting a proper $\eta$ value even harder. An internal mechanism that dynamically adapts the glow parameter according to the (possibly changing) external environment would therefore further enhance the autonomy of the PS agent. In Section \ref{secsec:ECM_eta} we will show how to implement this internal mechanism by means of adaptation through learning.

\section{Meta-learning within PS} \label{sec:ML_in_PS_design}
To enhance the PS model with a meta-learning component, we supplement the base-level clip-network (ECM) with additional networks, one for each meta-parameter $\xi$ (where $\xi$ could be, e.g.\ $\gamma$ or $\eta$). Each such meta-level network, which we denote by ECM$_{\xi}$ obeys the same principle structure and dynamic as the base-level ECM network as described in Section \ref{sec:PS_model}: it is composed of clips, its activation initiates a random-walk through the clips until an action-clip is hit, and its update rule is given by the update rule of Eq.~(\ref{eq:hupdate}), albeit with a different \emph{internal} reward: 
\be \label{eq:internal_update_rule}
	h_{\xi}^{(t+1)}(c_i,c_j) = h_{\xi}^{(t)}(c_i,c_j) - \gamma_{\xi} (h_{\xi}^{(t)}(c_i,c_j) - 1) + \lambda_{\xi}.~~~~
\ee

\begin{figure}[h]
\begin{center}
	\includegraphics[height=3.5cm]{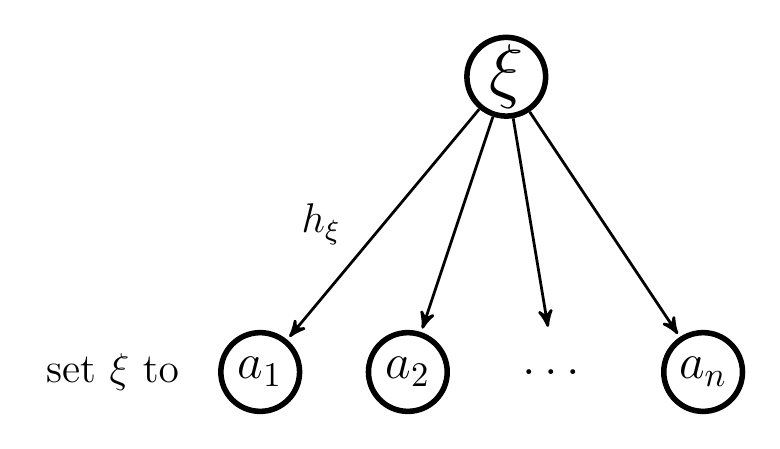}
\end{center}
	\caption{A schematic two-layered meta-level ECM$_{\xi}$ network, whose actions control the value of a general meta-parameter $\xi$.}
	\label{fig:xi_network}
\end{figure}

The meta-level ECM networks we consider in this work are two-layered, with a single percept-clip and several action-clips.  
The action clips of each meta-level network determine the next value of the corresponding meta-parameter. This is illustrated schematically in Fig.~\ref{fig:xi_network}. 

While the base-level ECM network is activated at every interaction with the environment (where each percept-action pair of the agent counts as a single interaction), a meta-level ECM$_{\xi}$ network is activated only every $\tau_{\xi}$ interactions with the environment. Following each activation, an action-clip is encountered and the meta-parameter $\xi$ (thus either $\gamma$ or $\eta$) is updated accordingly. At the end of each such $\tau_{\xi}$ time window the meta-level network receives an internal reward $\lambda_{\xi}$ which reflects how well the agent performed during the past $\tau_{\xi}$ interactions, or time steps, compared to the performance during the previous $\tau_{\xi}$ time window. This allows a statistical evaluation of the agent's performance in the last $\tau_{\xi}$ time window.

Specifically, we consider the quantity
\be 
	\label{eq:total_reward}
	\Lambda_{\xi}(T) = \sum_{t=T-\tau_{\xi}+1}^{T} \lambda^{(t)},  	
\ee
which accounts for the sum of rewards that the agent has received from the environment in the $\tau_{\xi}$ steps before the end of the time step $T$. The internal reward $\lambda_{\xi}$ is then set by comparing two successive values of such accumulative rewards:
\be
\label{eq:internal_reward}
\lambda_{\xi} = \mathop{\mathrm{sgn}}\left( \Delta_{\xi} \right)
\ee  
where $\Delta_{\xi}(T) = \frac{\Lambda_{\xi}(T) - \Lambda_{\xi}(T-\tau_{\xi})}{\max\{\Lambda_{\xi}(T),\Lambda_{\xi}(T-\tau_{\xi})\}}$ is the normalized difference in the agent's performance between two successive time windows, before time step $T$. In short, the meta-level ECM$_{\xi}$ network is rewarded positively (negatively) whenever the agent performs better (worse) in the latter time window (implementation insures that the corresponding $h$-values do not go below 1).  The normalization plays no role at this point, however the numerical value of $\Delta_{\xi}$ will matter later on.   
When there is no change in performance ($\Delta_{\xi}(T)=0$) the network is not rewarded.

The presented design requires the specification of several quantities for each meta-level ECM$_{\xi}$ network, including: the time window $\tau_{\xi}$, the number of its actions and the meaning of each action. In what follows we specify these choices for both the $\eta$ and the $\gamma$ meta-level networks. 

%We will see that internal parameters of different kinds induce different internal network, but that a single choice of these networks makes the agent highly flexible, in terms of the different environments with which it can cope. 

\subsection{The glow meta-level network (ECM$_{\eta}$) -- adaptation through learning only} \label{secsec:ECM_eta}
\begin{figure}[h]
\begin{center}
	\includegraphics[height=3.5cm]{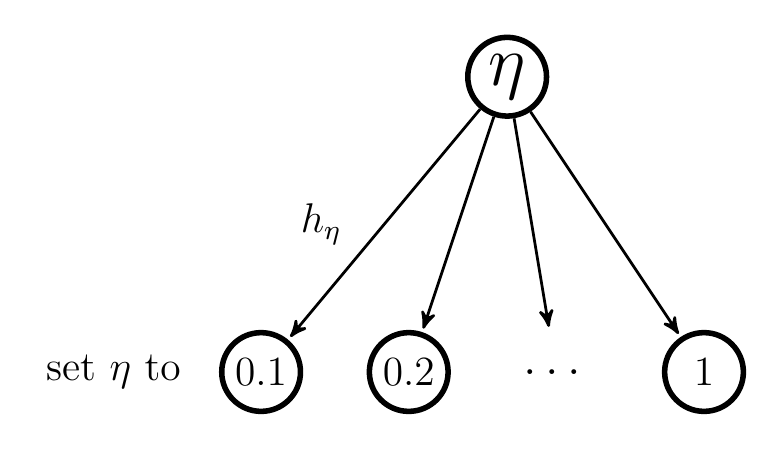}
\end{center}
	\caption{The glow meta-level network (ECM$_{\eta}$): The specific realization employed in this work.}
	\label{fig:eta_network}
\end{figure}

The glow meta-level network (ECM$_{\eta}$) we use in this work is depicted in Fig.~\ref{fig:eta_network}. The network is composed of a single percept ($S_{\eta}=1$) and $A_{\eta}=10$ actions which correspond to setting the $\eta$ parameter to one of 10 values from the set $\{0.1,0.2,\ldots,1\}$. 
The internal glow network is activated every $\tau_{\eta}$ times steps. This time window should be large enough so as to allow the agent to gather reliable statistics of its performance. It is therefore sensible to set $\tau_{\eta}$ to be of the order of the learning time of the agent, that is the time it takes the agent to reach a certain fraction of its asymptotic success probability (see also \cite{2013_Mautner_PSII}). The learning time of the PS was shown in \cite{2013_Mautner_PSII} to be linear in the number of percepts $S$ and actions $A$ in the base-level network. We thus set the time window to be $\tau_{\eta}=N_{\eta}SAS_{\eta}A_{\eta}$, which is also linear with the number of percepts $S_{\eta}$ and actions $A_{\eta}$ of the meta-level network. Here $N_{\eta}$ is a free parameter; the higher its value, the better the statistics the agent gathers. In this work, we set $N_{\eta}=30$ throughout, for all the examples we study.

%In order to change $\eta$ correctly by choosing out of this set the agent should gather statistics about the performance with the current $\eta$, i.e. it should be able to learn about the environment. Hence the time window $\tau_{\eta}$ should be of the order of the learning time, which is proportional to the number of percepts $S$ and the number of actions $A$ in the base-level network~\cite{2013_Mautner_PSII}, leading to the expression $\tau_{\eta}=N_{\eta}SAS_{\eta}A_{\eta}$ steps, where $N_{\eta}$ is a free parameter. The higher the value of $N_{\eta}$ parameter, the better the statistics that is gathered about the agent's performance. In this work, we set $N_{\eta}=30$ for all the considered examples.

The $h_{\eta}$-values of the $\eta$-network are updated through internal rewarding as described, and the PS agent learns with time what the preferable $\eta$ values in a given scenario are. 
The preferable $\eta$ values are further adjusted to account for changes in the environment. 
These continuous adjustments of the $\eta$-network then allow the PS agent to adapt to new environments by learning.

\subsection{The damping meta-level network (ECM$_{\gamma}$) -- combining reflexive adaptation with adaptation through learning} \label{secsec:ECM_gamma}
\begin{figure}[h]
\begin{center}
	\includegraphics[height=3.5cm]{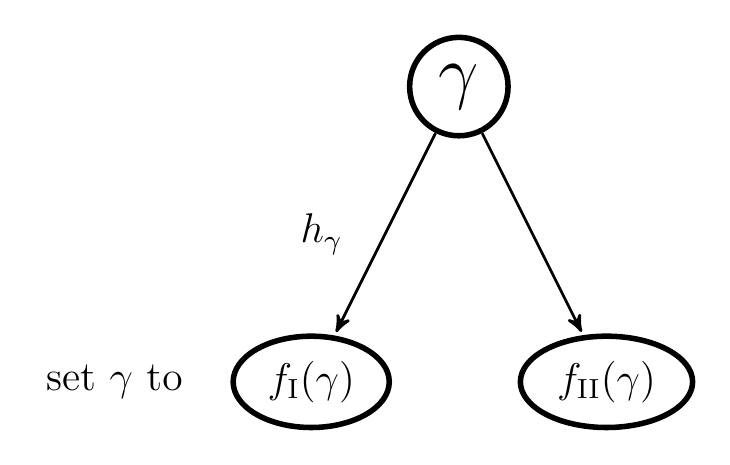}
\end{center}
	\caption{The damping meta-level network (ECM$_{\gamma}$): The specific realization employed in this work.}
	\label{fig:gamma_network}
\end{figure}

The second meta-learning network, the damping meta-level network (ECM$_{\gamma}$), is presented in Fig.~\ref{fig:gamma_network}. It is composed of only two actions which correspond to updating the $\gamma$ parameter by using one of two functions according to the following rules:

% var.Delta = var.Delta/max(var.success_avg, var.prev_success_avg);
% Delta = (Delta + DELTA_CONST)/(1+DELTA_CONST);
%net.gamma = (1-abs_Delta)*net.gamma + (abs_Delta-Delta)/2;
\be
\label{eq:gamma_rule1}
\text{Rule I:} \quad \gamma \leftarrow f_\mathrm{I}(\gamma) = (1-|\tilde{\Delta}_{\gamma}|)\gamma + \frac{|\tilde{\Delta}_{\gamma}|-\tilde{\Delta}_{\gamma}}{2}
\ee
and 
%net.gamma = (1-abs_Delta)*net.gamma + (abs_Delta+Delta)/2;
\be
\label{eq:gamma_rule2}
\text{Rule II:} \quad \gamma \leftarrow f_\mathrm{II}(\gamma) = (1-|\tilde{\Delta}_{\gamma}|)\gamma + \frac{|\tilde{\Delta}_{\gamma}|+\tilde{\Delta}_{\gamma}}{2}
\ee
where $\tilde{\Delta}_{\gamma} = \frac{\Delta_{\gamma}+C_{\gamma}}{1+C_{\gamma}}$ and $\Delta_{\gamma}$ is defined after Eq.~(\ref{eq:internal_reward}). Rule I invokes a reflexive increase of the $\gamma$ parameter when the agent's performance deteriorates, and a reflexive decrease when the performance improves. This rule (``natural rule'') is natural for typical RL scenarios: a drop of performance is assumed to signify a change in the environment, at which point the agent should do well to forget what it learned thus far, and focus on exploring new options - in the PS both are achieved by the increase of $\gamma$. In contrast, if the environment is in a stable phase, as the agent learns, the performance improves, causing $\gamma$ to decrease, which will lead to optimal performance. 
Rule II (``opposite rule'') is chosen to do exactly the opposite, namely performance increase causes the agent to forget. Our main purpose for the introduction of this rule is to demonstrate the flexibility of the meta-learning agent to learn even the correct strategy of updating $\gamma$. Although in all the environments which are typically considered in literature, and in this work, the natural rule is the better choice, and thus could in principle be hard-wired, our agent is challenged to learn even this\footnote{It is possible to concoct settings where the opposite rule may be beneficial using minor and major rewards. The environment may use minor rewards to train the agent to a deterministic behavior over certain time periods, after which a major reward (dominating the total of all small rewards) is issued only if the agent nonetheless produced random outcomes all along. If the periods are appropriately tailored, this can train the meta-learning network to prefer the opposite rule. The study of such pathological settings are not of our principal interest in this work.}.
%Rule II does the opposite, namely the PS agent forgets when it performs well and remembers more when it underperforms.
The role of $C_{\gamma}$ parameter, which we set to $C_{\gamma}=0.2$ throughout this work, is to avoid unwanted increase of $\gamma$ under statistical fluctuations. Note that the functions in Eqs.~(\ref{eq:gamma_rule1}) and~(\ref{eq:gamma_rule2}) ensure that $\gamma\in[0,1]$.

The described $\gamma$-network is activated every $\tau_{\gamma} = N_{\gamma}\tau_{\eta}$ steps, where $\tau_{\eta}$ is the time window of the glow network as defined above, and where $N_{\gamma}$ is a free parameter of the $\gamma$-network, which we set it to $N_{\gamma}=5$ throughout the paper. The agent first learns an estimate of the range of an optimal $\eta,$ and changes $\gamma$ afterwards. This is assured by choosing the time window for the $\gamma$-network larger than for the $\eta$-network.
This relationship between the time windows $\tau_{\gamma}$ and $\tau_{\eta}$ is required in order for the PS agent to gain a meaningful statistics during $\tau_{\gamma}$ steps. 
Otherwise, if $\eta$ is not learned first, the agent's performance will significantly fluctuate, leading to erratic changes of $\gamma$ through the reflexive adaptation rule. Note that large fluctuations in $\gamma$ yield very poor learning results as even moderate values of $\gamma$ lead to a rapid forgetting of the agent.

%in order for the PS agent to gain knowledge about a range of preferable $\eta$ parameters that will mostly lead to an improvement in the agents performance and make it possible to intentionally change the $\gamma$ parameter by comparing two last distinct statistics gained during wider time windows. Without the knowledge about desirable $\eta$ parameters the agents performance will fluctuate a lot leading to a reflective random change of $\gamma$.

The meta-learning by the ECM$_{\gamma}$ network is realized as follows. Starting from an initially random value, the $\gamma$ parameter is adapted both via direct learning in the $\gamma$-network and via reflexive adaptation through rule I or rule II.
Given that the overall structure of the environment was learned (\emph{i.e.} whether the natural rule I or opposite rule II is preferable), $\gamma$ is henceforth adapted reflexively. These reflexive rules reflect an a-priory knowledge about what strategy is preferable in given environments.
We note that the $\gamma$ parameters could be learned without such reflexive rules, by using networks which directly select the $\gamma$ values (like in the case of the $\eta$ network), however such approaches have shown to be much less efficient. In general, reflexive adaptation of the meta-parameters is preferable to adaptation through learning as it is simpler. The need for learning adaptation arises when the landscape of optimal values of the meta-parameters is not straightforward, as is the case for the $\eta$ parameter, as illustrated in Fig.~\ref{fig:nship_eta_reconstructed}.
	
\section{Simulations} \label{sec:results}
To examine the proposed meta-learning mechanism we next evaluate the performance of the meta-learning PS agent in several environments, namely the invasion game, the $n$-ship game and variants of the grid-world setting. These environments were chosen because of their different structures, which exhibit different optimal damping and glow parameters. The goal is that the PS agent will adjust its meta-parameters properly, so as to cope well with these different tasks. To challenge the agent even further, each of the three environments will suddenly change, thereby forcing the agent to readjust its parameters accordingly. Critically, for all tasks the same meta-level networks are used, along with the same choice of free parameters ($N_{\eta}=30$, $N_{\gamma}=5$, and $C_{\gamma}=0.2$), as described in Sections \ref{secsec:ECM_eta}-\ref{secsec:ECM_gamma}.

To demonstrate the role of the meta-learning mechanism, we compare the performance of the meta-learning PS agent to the performance of the PS agent without this mechanism. Without the meta-learning the PS agent starts the task with random $\gamma$ and $\eta$ parameters and does not change them afterwards. To show the importance of learning the optimal $\eta$ parameter (which may not be as obvious as for the case of $\gamma$) we construct a second reference PS agent for comparison, which uses the $\gamma$-network to adjust the $\gamma$ parameter, but takes a random choice out of the possible $\eta$-actions in the $\eta$-network.

%The performance of the PS agents in the invasion game and the n-ship game is evaluated by 
%is evaluated differently in the tasks we consider. In the 
%by obtaining the probability of the correct actions of a single agent and averaging over several agents afterwords. This is done when the central ECM is relatively small, e.g. in the invasion game and the n-ship game, the probability of the correct action of each agent is extracted individually from its ECM network. However, when the ECM is larger, e.g. in the grid-world, calculating the probability of each possible sequence of actions becomes more involved. 
%We therefore calculate the average performance by
%as the fraction of correct actions done by several agents at a certain time step.

\subsection{The invasion game}
We start with the simplest task: the invasion game (see Section \ref{secsec:damping_parameter}). As before, the agent is rewarded with $\lambda=1$ whenever it manages to block the attacker and it has to learn whether the attacker will go left or right, after presenting one of two symbols. We consider, once again, the scenario in which the attacker switches between two strategies every fixed number of trials. In one phase of the game it goes left (right) after showing a left (right) symbol, whereas in the other phase it does the opposite. This is repeated several times. The task of the agent is to block the attacker regardless of its strategy. We recall that in such a scenario (see Fig.~\ref{fig:invasion_game_fixed_change}) the basic PS agent with fixed meta-parameters ($\gamma=0, \eta=1$) can only cope with the first phase, but fails completely at the second.

\begin{figure}[H]
	%\begin{center}						
		\includegraphics[width=9cm]{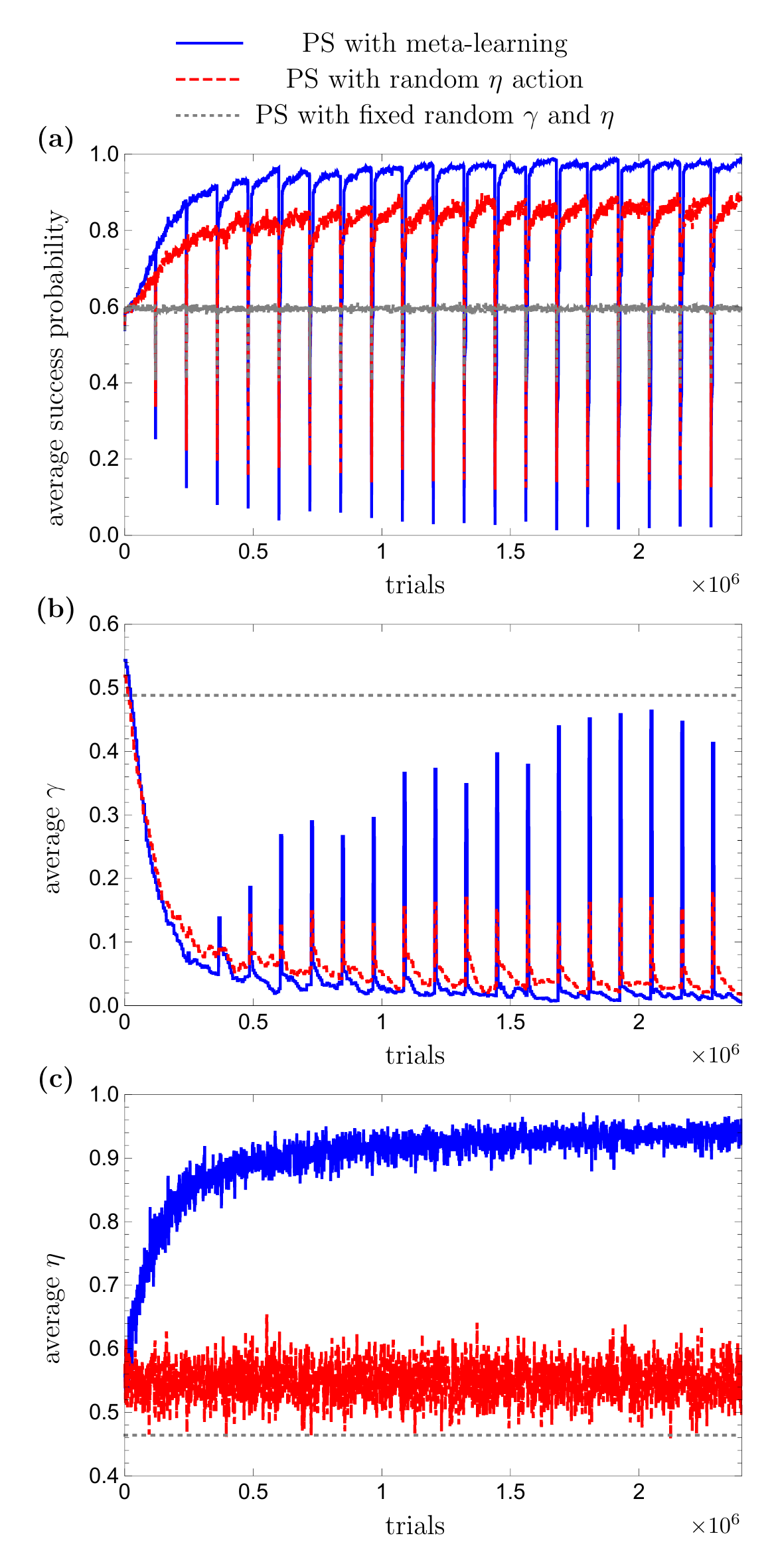}
	%\end{center}
	\caption{(Color online) \emph{Invasion game:} The attacker inverts its strategy every $1.2\times 10^5$ steps. Three types of PS agents are depicted: with full meta-learning capability (in solid blue), with adjusted $\gamma$ value but with $\eta$ value that is chosen randomly from the ECM$_{\eta}$ network (in dashed red), and agents whose $\gamma$ and $\eta$ values are fixed to random values, each agent with its own values (in dotted gray). \textbf{Top:} The performances of the different agents are shown as a function of trials; \textbf{Middle:} The average $\gamma$ value is shown as a function of trials; \textbf{Bottom:} The average $\eta$ value is shown as a function of trials. The simulations were done by averaging over 100 agents, where for each agent the success probabilities were extracted directly from its base-level ECM network.}
	\label{fig:invasion_game_ML}
\end{figure}

Fig.~\ref{fig:invasion_game_ML} (a) shows in solid blue the performance of the meta-learning PS agent, in terms of average success probabilities, in this changing invasion game. Here each phase lasts $1.2 \times 10^5$ steps, and the attacker changes its strategy 20 times. It is seen that with time the average success probability of the PS agents increases towards optimal values and that the agents manage to block the attacker equally well for both of its strategies. This performance is achieved due to meta-learning of the $\gamma$ and $\eta$ parameters,  the dynamics of which are shown in solid blue in Fig.~\ref{fig:invasion_game_ML} (b) and (c), respectively. 
It is seen in Fig.~\ref{fig:invasion_game_ML} (b) that after some time and several phase changes, the value of the $\gamma$ parameter raises sharply whenever the attacker changes its strategy, and decays toward zero during the following phase. 
This allows the agent to rapidly forget its knowledge of the previous strategy and then to learn the new one. Fig.~\ref{fig:invasion_game_ML} (c) shows the $\eta$ parameter dynamics. As explained in Section \ref{secsec:glow_parameter} the optimal glow value for the invasion-game is $\eta=1$, as the environment induces no temporal correlations between previous actions and rewards. The meta-learning agent begins with an

\noindent $\eta$-network that has a uniform action probability. Yet, with time, its meta-level ECM$_{\eta}$ network learns and the average $\eta$ parameter approaches the optimal value of $\eta=1$.

To show the advantage of the meta-learning networks we next consider the performance of agents without this mechanism. First, we look at the performance of a PS agent with fixed random $\gamma$ and $\eta$ parameters as shown in Fig.~\ref{fig:invasion_game_ML} (a) in dotted gray. It is seen that on average, such an agent performs rather poor, with an average success rate of $0.6$. This can be expected, as most of the $\gamma$ and $\eta$ values are in fact harmful for the agent's success. The average value of each parameter goes to $0.5$ as depicted in Fig.~\ref{fig:invasion_game_ML} (b) for the $\gamma$ parameter and in Fig.~\ref{fig:invasion_game_ML} (c) for the $\eta$ parameter in dotted gray (the slight deviation from $0.5$ is due to finite sample size).

A more challenging comparison is shown in Fig.~\ref{fig:invasion_game_ML} (a) in dashed red, where the agent adjusts its $\gamma$ parameter exactly like the meta-learning one, but uses an $\eta$-network (the same one as the meta-learning agent) that does not learn or update. It is seen that such an intermediate agent can already learn both phases to some extent, but does not reach optimal values. This is because small $\eta$ values -- corresponding to sustained glow over several learning cycles -- are harmful in this case. The dynamics of the parameters of this PS agent are shown in Fig.~\ref{fig:invasion_game_ML} (b) and (c) in dashed red, where $\gamma$ behavior is similar to the one of the meta-learning agent, and the average $\eta$ fluctuates around $\eta=0.55$, which is the average value of the $10$ possible actions in the $\eta$-network.

\begin{figure}[h]
	\begin{center}
		\begin{minipage}{9cm}
			\includegraphics[width=8cm]{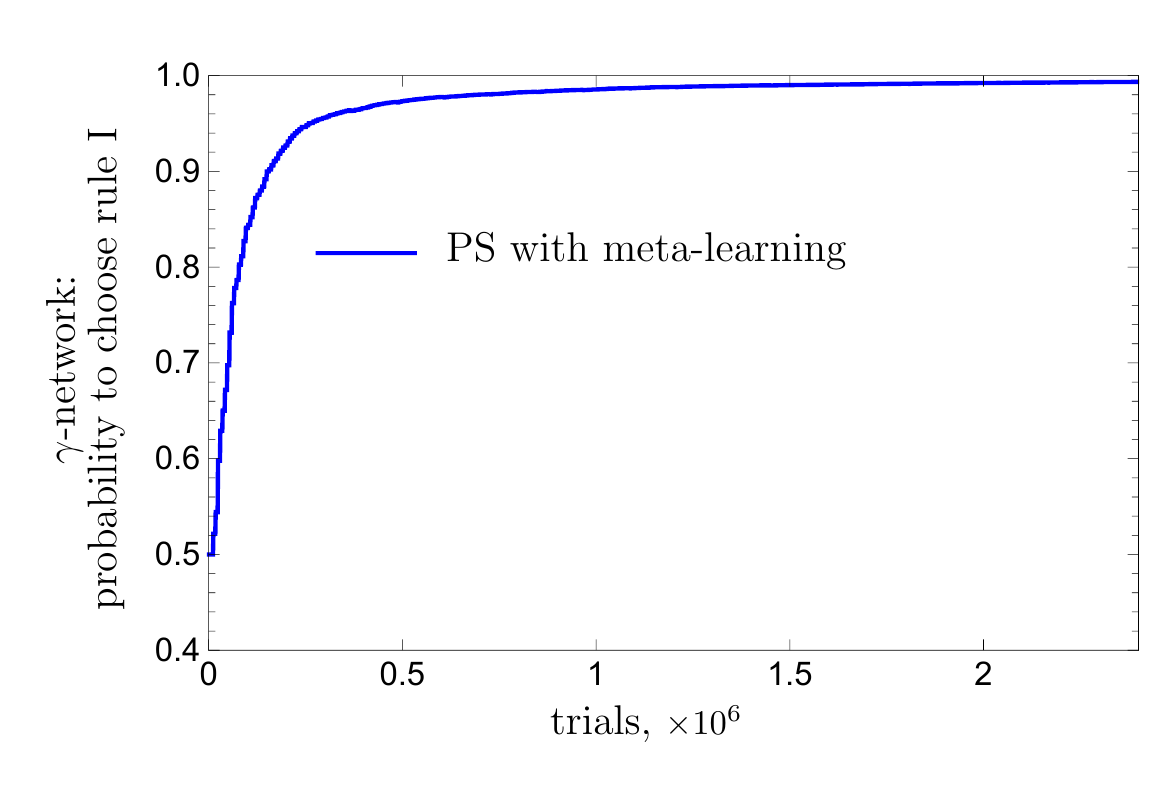}
		\end{minipage}
	\end{center}
	\caption{\emph{Invasion game:} The attacker inverts its strategy every $1.2\times 10^5$ steps as in Fig.~\ref{fig:invasion_game_ML}. The performance of the $\gamma$-network of the meta-learning PS is shown as a function of trials, in terms of the probability to choose rule I (see Eq.~\ref{eq:gamma_rule1}). The simulation was done by averaging over 100 agents.}
	\label{fig:invasion_game_full_ML-gamma}
\end{figure}

In this example we encounter for the first time the trade-off between flexibility and learning time. The meta-learning agent exhibits high flexibility and robustness, as it manages to repeatedly adapt to changes in the environment. However, this comes with a price: the learning time of the agent slows down and the agent requires millions of trials to master this task. 
This is, however, to be expected. Not only that the agent has to learn how to act in a changing environment, but it must also learn how to properly adapt its meta-parameters, and the latter occurs at the time-scales of $\tau_\gamma = 6000$ elementary cycles. The agent begins with no bias whatsoever regarding its a action pattern or its $\gamma$ and $\eta$ parameters. Furthermore, the agent begins with no a-priory knowledge regarding the inherent nature of the rewarding process of the environment: is it a typical environment (where the agent should prefer the natural rule), or is it an untypical environment which ultimately rewards random behavior (where the opposite rule will do better)? This too needs to be learned. Fig.~\ref{fig:invasion_game_full_ML-gamma} shows the average probability of choosing rule I (Eq.~(\ref{eq:gamma_rule1})) in the $\gamma-$network as a function of trials. It is seen that with time the $\gamma-$network chooses to update the $\gamma$ parameter according to rule I with increasing probability, reflecting the fact that in this setup the environment acts indeed according to the natural rule.

\subsection{The $n$-ship game} \label{secsec:n-ship_game}
In the $n$-ship game (see Section~\ref{secsec:glow_parameter}) the environment rewards the agent depending on its previous actions. In what follows we consider a dynamic $n$-ship game, that is we allow $n$ to change with time. In particular, the environment starts with $n=1$ (where no temporal correlations exist) and increases the number of ships $n$ by one, every $3.5 \times 10^5 \times n$ steps. As explained in Section~\ref{secsec:glow_parameter} each $n$-value requires a different glow parameter $\eta$. This scenario therefore poses the challenge of continuously adjusting the glow parameter.   

\begin{figure}[H]
	\begin{center}
		\begin{minipage}{9cm}
			\includegraphics[width=9cm]{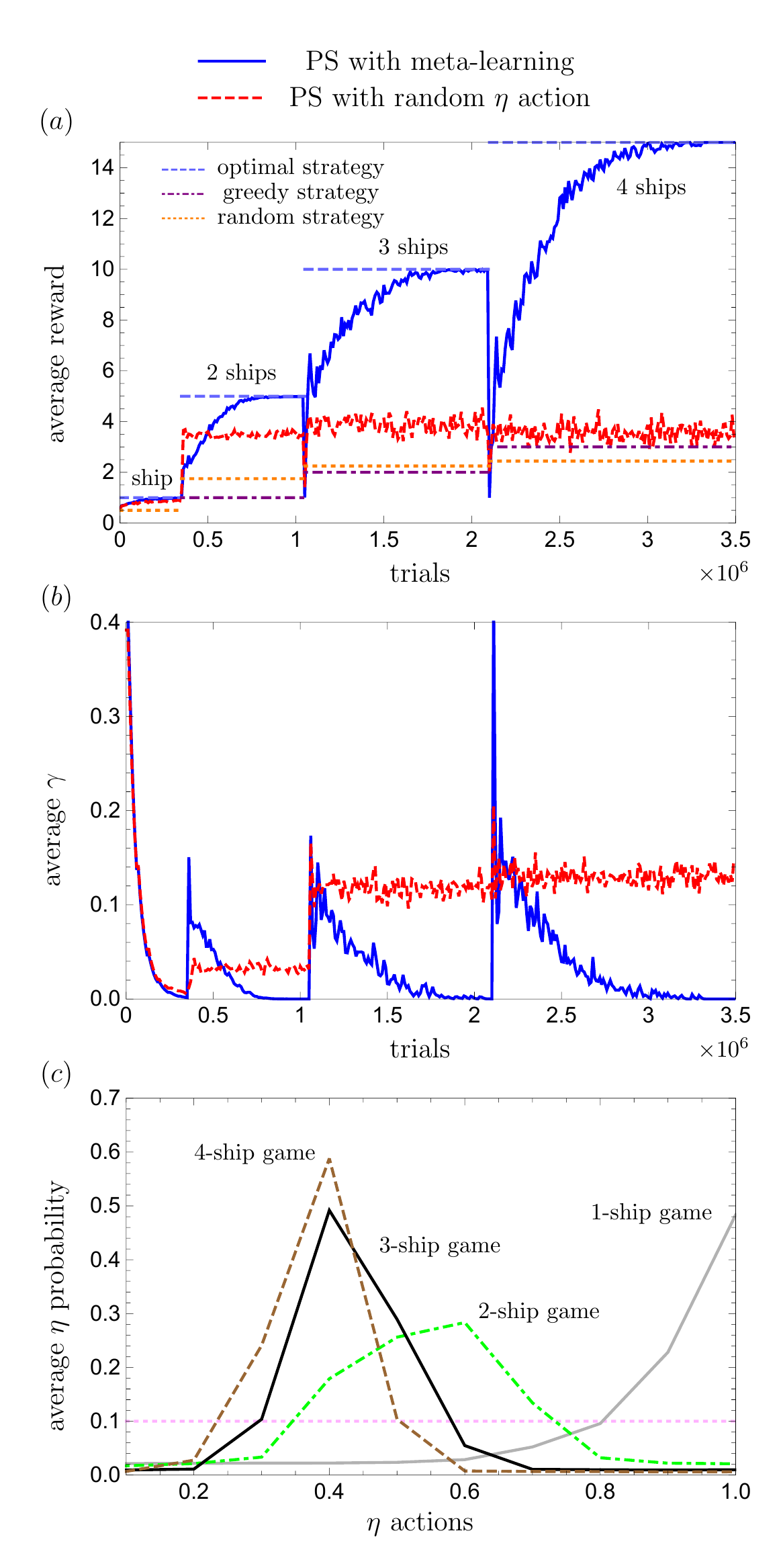}
		\end{minipage}
	\end{center}
\caption{(Color online) \emph{n-ship game:} The number of ships $n$ increases from one to four. Each phase lasts for  $3.5 \times 10^5 \times n$ trials. Two types of PS agents are depicted: with full meta-learning capability (in solid blue), and with adjusted $\gamma$ value but with $\eta$ value that is chosen randomly from the ECM$_{\eta}$ network (in dashed red). \textbf{Top:} The performance of the two different agents is shown as a function of trials in terms of average reward. For each phase the average reward of the optimal strategy, a greedy strategy and a fully random one is plotted in dashed light-blue, dotted-dashed purple, and dotted orange, respectively; \textbf{Middle:} The average $\gamma$ values of the two different kinds of agents are shown as a function of trials; \textbf{Bottom:} For the meta-learning PS agent the probability to choose each of the 10 $\eta$-actions is plotted at the end of each phase in a different plot. Connecting lines are shown to guide the eyes. The simulations were done by averaging over 100 agents, where for each agent the average reward was extracted directly from its base-level ECM network.}
	\label{fig:n-ship_game_ML}
\end{figure}

Fig.~\ref{fig:n-ship_game_ML} (a) shows in solid blue the performance of the meta-learning PS agent in this changing $n$-ship game. The best possible reward is indicated by a dashed blue horizontal line, and it is seen that such agents learn to perform optimally, for all number of ships $n$. This success is made possible by the meta-learning mechanism. 

First, the $\gamma$ parameter is adjusted, such that the agent forgets whenever its performance decrease and vice versa (see Eq.~(\ref{eq:gamma_rule1})). Here we assume that the $\gamma$-network already learned in previous stages that the environment follows the natural rule (we used an h-value ratio of $10^5$ to $1$ for rule I). This $\gamma$-network leads to a dynamics of the average $\gamma$ parameter shown in Fig.~\ref{fig:n-ship_game_ML}(b) in solid blue. It is seen that whenever the environment changes, the $\gamma$ parameter increases, thereby allowing the agent to forget its previous knowledge. A slow decrease of the damping parameter makes it then possible for the agent to learn how to act in the new setup. 

Second, the glow parameter $\eta$ is adjusted dynamically. Fig.~\ref{fig:n-ship_game_ML} (c) shows the probability distribution of choosing each action of the $\eta$-network at the end of each phase. It is seen that as $n$ grows, the $\eta$-network learns to choose a smaller and smaller glow parameter value, which allows the back propagation of the reward from the final ship to the first $n-1$ ships.
A similar trend was observed in Fig.~\ref{fig:nship_eta_reconstructed} where larger $n$ values result with smaller

\noindent optimal $\eta$ values. As shown in Fig.~\ref{fig:n-ship_game_ML} (c) the meta-learning PS agent essentially captures the same knowledge in its $\eta$-network. Yet, this time the knowledge is obtained through the agent's experience, rather than by an external party.

The PS agent without the meta-learning is not able to achieve similar performance. Performance of an agent with fixed random $\gamma$ and $\eta$ is poor and not shown, because its behavior was close to a random strategy (dotted orange horizontal lines in Fig.~\ref{fig:n-ship_game_ML} (a)). This performance is expected, because most of the values are harmful for the agent's success. We only show the more challenging comparison (dashed red in Fig.~\ref{fig:n-ship_game_ML}), where the agent adjusts its $\gamma$ parameter exactly like the meta-learning one, but uses an $\eta$-network (the same one as the meta-learning agent) which does not learn or update. It is seen that for $n=1$ such an intermediate agent can cope with the environment, but that for higher values of $n$ it fails to reach the optimal performance because of a random $\eta$ parameter, and achieves only a mixture of a greedy strategy (dot-dashed purple horizontal lines) and an optimal strategy (dashed light blue horizontal lines).

\subsection{The grid-world task} \label{secsec:grid-world}
\begin{figure*}[ht!]
	\centering
		%\begin{minipage}{9cm}						
\includegraphics[width=7in]{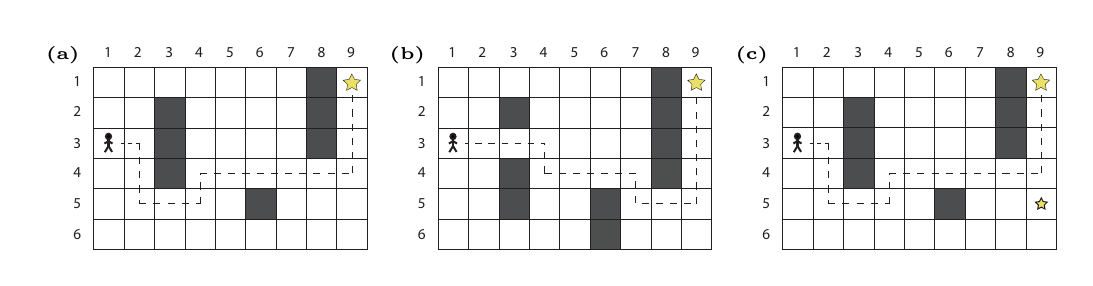}
			%\end{minipage}
	\caption{(Color online) Three setups of the grid-world task. \textbf{Left:} The basic grid-world as presented in Ref.~\cite{sutton1990integrated}; \textbf{Middle:} Some of the walls are positioned differently; \textbf{Right:} The basic grid-world with a distracting small reward $\lambda_{\min}$ placed 12 steps from the agent. In all three setups a large reward of $\lambda_{\max}$ awaits the agent in 14 steps.}
	\label{fig:grid_world_3_phases}
\end{figure*}

As a last example, we consider a benchmark problem in the form of the grid-world setup as presented in Ref.\ \cite{sutton1990integrated}. This is a delayed-reward scenario, where the agent walks through a maze and gets rewarded with $\lambda=1$ only when it reaches its goal. At each position, the agent can move in one of four directions: left, right, up, or down. Each move counts as a single step. Reaching the goal marks the end of the current trial and the agent is then reset to its initial place, to start another round. The basic PS agent was shown to perform well in this benchmark task \cite{2014_Melnikov_PSIII}. Here, to challenge the new meta-learning scheme we situate the agent in three different kinds of grid-worlds: (a) The basic grid-world of Ref. \cite{sutton1990integrated}, illustrated in the left part of Fig.~\ref{fig:grid_world_3_phases}; (b) The same sized grid-world with some of the walls positioned differently, as shown in the middle part of Fig.~\ref{fig:grid_world_3_phases}; and (c) The original grid-world, but with an additional small distracting reward $\lambda_{\min}=\frac{1}{3}$, placed only 12 steps from the agent, shown in the right part of Fig.~\ref{fig:grid_world_3_phases}. The game then ends either when the big reward $\lambda_{\max}=1$ or the small reward $\lambda_{\min}$ are reached. In all cases the shortest path to the (large) reward is composed of 14 steps. 

Since it is the same agent that goes through each of the phases, it has to forget its previous knowledge whenever a new phase is encountered, and to adjust its meta-parameters to fit the new scenario. The third phase poses an additional challenge for the agent: for optimal performance, it must avoid taking the small reward and aim at the larger one. 

\begin{figure}[H]
	\begin{center}
		\begin{minipage}{9cm}
			\includegraphics[width=9cm]{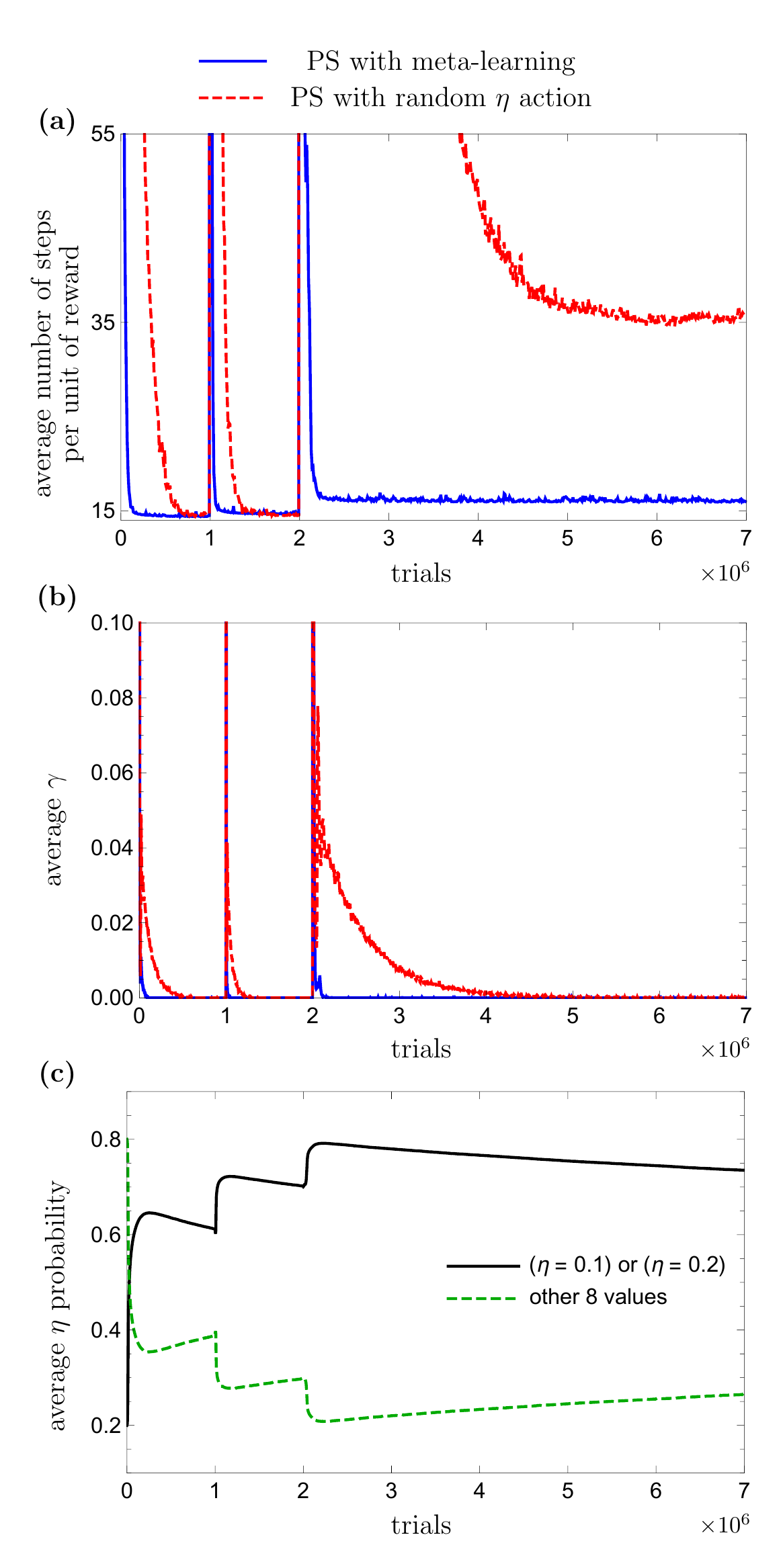}
		\end{minipage}
		\end{center}
	\caption{(Color online) \emph{Grid-world task:} Two types of PS agents are depicted: with full meta-learning capability (in solid blue), and with adjusted $\gamma$ value but with $\eta$ value that is chosen randomly from the ECM$_{\eta}$ network (in dashed red). \textbf{Top:} The performances of the two different agents are shown as a function of trials in terms of average number of steps per unit reward; \textbf{Middle:} The average $\gamma$ values of the two different kinds of agents are shown as a function of trials; \textbf{Bottom:} For the meta-learning PS agent the probability to choose $\eta=0.1$ or $\eta=0.2$ and the probability to choose either of the other 8 $\eta$-actions are plotted as a function of trials; The first two phases of the game last $10^6$ trials, whereas the last phase lasts $5\times 10^6$ trials. These phases correspond to three different kinds of grid-worlds shown in Fig.~\ref{fig:grid_world_3_phases}. The simulations were done by averaging over $10^4$ agents.}
	\label{fig:grid_world_performance}
\end{figure}

Fig.~\ref{fig:grid_world_performance} (a) shows in solid blue the performance of the meta-learning PS agent throughout the three different phases of the grid-world. The performance is shown in 

\noindent terms of steps per reward as a function of trials. In all cases the optimal performance is 14 steps per one reward. In the last phase, a greedy agent would reach the small reward of $\lambda_{\min}=\frac{1}{3}$ in 12 steps, thus resulting with 36 steps per unit of reward. It is seen that the meta-learning agent performs optimally in all phases, except of the last phase, where the performance is only suboptimal with an average of about $16$ steps per unit reward (instead of $14$). This flexibility through all phases is achieved due to the adjustments of the $\gamma$ parameter, whose progress over time is shown in Fig.~\ref{fig:grid_world_performance} (b) in solid blue. Similar to the $n$-ship game, we assume that the $\gamma$-network has already learned that the environment uses the ``straightforward" logic, by setting the $h$-value ratio of $10^5$ to 1 for choosing rule I. 
The PS agent with the same $\gamma$-network, but without updated $\eta$-network, performs similarly in the first two phases (Fig.~\ref{fig:grid_world_performance} (a) in dashed red) in terms of finding eventually an optimal path. This is to be expected because for finding an optimal path it is only necessary that $0 < \eta < 1$.
It is seen, however, that this agent learns much slower than the full meta-learning agent, so that hundreds of thousands more steps are required on average to find an optimal path. This is also reflected in the behavior of the $\gamma$ parameter: with a random $\eta$ value, the $\gamma$ parameter goes to zero much slower, as shown in Fig.~\ref{fig:grid_world_performance} (b).

The importance of the $\eta$ parameter is however better demonstrated in the third phase, where the difference between the achieved performance of the agent with and without $\eta$-learning is very significant. In particular, the PS agent with a random $\eta$ converges to the greedy strategy and gets a unit of reward every $36$ steps (Fig.~\ref{fig:grid_world_performance} (a) in dashed red). 
The reason is that optimal performance (a unit of a reward every $14$ steps) can only be achieved by setting the $\eta$ parameter to a value from a certain, limited, range, which we analyze next.

The range of optimal $\eta$ values can be obtained by focusing on the (4, 9) location in the grid-world (see Fig.~\ref{fig:grid_world_3_phases} (c)). In this location, the agent has two possible actions that lead faster to the large and small rewards, namely up and down, respectively. Because these two actions result with a faster reward, edges corresponding to these actions are enhanced stronger than for the other actions. This enhancement is gained by adding the increments $g^{(t)}(c_i,c_j)\lambda$ to the $h$-values at the end of each game, as one can see from the update rule in Eq.~(\ref{eq:hupdate2}). If the PS agent follows the greedy strategy, then this increment is equal to $\lambda_{\min}=1/3$ and is added to the edge corresponding to the action ``down". For the optimal strategy the increment is $\lambda_{\max}(1-\eta)^2 = (1-\eta)^2$ (since $\lambda_{\max}=1$), because the large reward occurs two decisions away from the current position and the $g$-value is damped from the value of $1$ to the value of $(1-\eta)^2$. The optimal strategy will prevail only if the increment in each game is larger than for the greedy strategy. This is the only case when $0<\eta< 1-\sqrt{1/3} < 0.43$. Most of the actions in the $\eta$-network of the PS agent have values larger than $0.43$, therefore the agent with random $\eta$ actions converges to the greedy strategy and does not get the best possible reward. The PS agent with meta-learning is able to learn to use the $\eta$ parameter from the optimal range, and as shown in Fig.~\ref{fig:grid_world_performance} (c) the agent indeed mostly uses the values of $\eta=0.1$ and $0.2$. 

\section{Summary and Discussion} \label{sec:discussion}
We have developed a meta-learning machinery that allows the PS agent to dynamically adjust its own meta-parameters.
This was shown to be desirable by demonstrating that, like in other AI schemes, no unique choice of the model's learning parameters can account for all possible tasks, as optimal values for the meta-parameters vary from one task environment to another. 
We emphasize that the presented meta-learning component is based on the same design as the basic PS, using random walk on clip-networks as the central information processing step. It is therefore naturally integrated into the PS learning framework, preserving the model's stochastic nature, along with its simplicity.

The basic PS has two principal meta-parameters: the damping parameter $\gamma$ and the glow parameter $\eta$. For each meta-parameter we have assigned a meta-learning clip-network, whose actions control the parameter's value. Each meta-level network is activated every fixed number of interactions with the environment. 
This time 

\noindent window allows the agent to gather statistics about its performance, so as to monitor its recent success rates and thereby to evaluate the setting of the corresponding parameters. 
When the agent's success increases, the previous action of the meta-level network is rewarded positively, otherwise, when the performance deteriorates, a negative reward is assigned (no reward is assigned when there is no change in the agent's success). As a result, the probability that the random walk on the meta-level network hits more favorable action clips increases with time and the meta-level network essentially learns how to properly adjust the corresponding parameter in the current environment.

The meta-learning process occurs on a much larger time-scale compared to the base-level network learning time scale. This is necessary as meta-level learning requires statistical knowledge of the agent's performance, which is directly controlled by the base-level network, whose learning time is linear with the state space of the task, represented by the number of percepts and actions in the base-level network.

In meta-level learning we have distinguished between adaptation through learning, which exploits the entire individual history of the agent to update the value of the meta-parameter, and reflexive adaptation, which updates the meta-parameter using only recent, localized information of the agent's performance. We saw that the glow parameter can be well adjusted with a full learning network that is only via adaptation through learning, whereas for the damping parameter, it is more sensible to combine the two kinds of adaptations.

The presented meta-learning scheme was examined in three different environmental scenarios, each of which requires a different set of meta-parameters for optimal performance. 
Specifically, we have considered the ``invasion game", where there are no temporal correlations between actions and rewards (implying that the optimal glow value is $\eta_{\text{opt}}=1$), the ``$n$-ship game" where temporal correlations do exist and $\eta_{\text{opt}}$ depends on $n$, and finally the ``grid-world", a real-world scenario with delayed rewards, for which it is sufficient that $\eta_{\text{opt}} \neq 1$ in the basic setup, but requires that $\eta_{\text{opt}} << 1$ in the more advanced setup, where the agent can be distracted by a small reward. 

In all scenarios, the environment furthermore suddenly changes, thus requiring the agent to also adjust its forgetting parameter $\gamma$.
% so as to adapt to the new environment. 
Overall, situating an agent in such changing environments enforces it to repeatedly and dynamically revise its internal settings. The meta-learning PS agent was shown to cope well in all scenarios, reaching success probabilities that approach near-optimal or optimal values.

% comparison
For comparison, we checked how a PS agent with fixed set of random meta-parameters would handle these scenarios, and observed that such an agent would perform significantly worse. This is not surprising, as most of the possible meta-parameter values (especially those of the $\gamma$ parameter) are harmful for the agent. Therefore, for a more challenging comparison, we checked the performance of an agent that adapts its forgetting parameter $\gamma$ in exactly the same way as the meta-learning agent, but chooses its glow parameter $\eta$ randomly, out of the same set of actions available in the $\eta$-network we used.
% This was done to verify specifically that the learning of the $\eta$-network has a meaningful role. 
Such an intermediate agent performed better than the basic PS agent with random choice of meta-parameters, but substantially worse than the full meta-learning agent. This demonstrates the importance of adjusting both $\gamma$ and $\eta$ in a proper way. In particular, it shows that the learning of the $\eta$-network plays a crucial role.  

% additional parameters
Importantly, throughout the paper, we used the same set of choices for the meta-learning scheme. In particular, we used the same meta-level networks ECM$_{\gamma}$ (including the reflexive rules of the $\gamma$-parameter adaptation) and ECM$_{\eta}$, and the same time windows $\tau_{\gamma}$ and $\tau_{\eta}$. This indicates that the suggested meta-learning scheme is robust, as it requires no further adjustment of additional parameters by an external party,  for all the cases we have considered.

\bibliography{bibliography}
\end{document}